\documentclass[lettersize,journal]{IEEEtran}
\usepackage{amsmath,amsfonts}
\usepackage{algorithmic}
\usepackage{algorithm}
\usepackage{array}
\usepackage[caption=false,font=normalsize,labelfont=sf,textfont=sf]{subfig}
\usepackage{textcomp}
\usepackage{stfloats}
\usepackage{url}
\usepackage{verbatim}
\usepackage{graphicx}
\usepackage{cite}
\usepackage{comment}
\usepackage{color}

\newcommand{\lzf}[1]{{\color{black} #1}}

\newcommand{\sysName}{{Sketch2Human}}

\newcommand{\eg}{\textit{e}.\textit{g}.}

\begin{document}

\title{\sysName: Deep Human Generation with Disentangled Geometry and Appearance Control}

\author{
Linzi Qu, Jiaxiang Shang, Hui Ye, Xiaoguang Han, and Hongbo Fu

\IEEEcompsocitemizethanks{\IEEEcompsocthanksitem 
Corresponding author: Hongbo Fu \\
L. Qu, H. Ye, and H. Fu are with the School of Creative Media, City University of Hong Kong. 
E-mail: linziqu2-c@my.cityu.edu.hk, huiye4@cityu.edu.hk, hongbofu@cityu.edu.hk

\IEEEcompsocthanksitem J. Shang is with the Department of Computer Science \& Engineering, HKUST. E-mail: jshang@cse.ust.hk

\IEEEcompsocthanksitem X. Han is with Shenzhen Research Institute of Big Data, Chinese University of Hong Kong, Shenzhen. E-mail: hanxiaoguang@cuhk.edu.cn}
}

\markboth{Journal of \LaTeX\ Class Files,~Vol.~14, No.~8, August~2021}%
{Shell \MakeLowercase{\textit{et al.}}: A Sample Article Using IEEEtran.cls for IEEE Journals}


\maketitle

\begin{abstract}
Geometry- and appearance-controlled full-body human image generation is an interesting but challenging task. Existing solutions are either unconditional or dependent on coarse conditions (e.g., pose, text), thus lacking explicit geometry and appearance control of body and garment.  
Sketching offers such editing ability and has been adopted in various sketch-based face generation and editing solutions. However, directly adapting sketch-based face generation to full-body generation often fails to produce high-fidelity and diverse results due to the high complexity and diversity in the pose, body shape, and garment shape and texture. Recent geometrically controllable diffusion-based methods mainly rely on prompts to generate appearance and it is hard to balance the realism and the faithfulness of their results to the sketch when the input is coarse. This work presents \sysName, the first system for controllable full-body human image generation guided by a semantic sketch (for geometry control) and a reference image (for appearance control). Our solution is based on the latent space of StyleGAN-Human with inverted geometry and appearance latent codes as input. Specifically, we present a sketch encoder trained with a large synthetic dataset sampled from StyleGAN-Human's latent space and directly supervised by sketches rather than real images. Considering the entangled information of partial geometry and texture in StyleGAN-Human and the absence of disentangled datasets, we design a novel training scheme that creates geometry-preserved and appearance-transferred training data to tune a generator to achieve disentangled geometry and appearance control. Although our method is trained with synthetic data, it can handle hand-drawn sketches as well. 
Qualitative and quantitative evaluations demonstrate the superior performance of our method to state-of-the-art methods. We will release the code upon the acceptance of the paper.
\end{abstract}


\section{Introduction}\label{sec:intro}
\begin{figure*}
  \includegraphics[width=\textwidth]{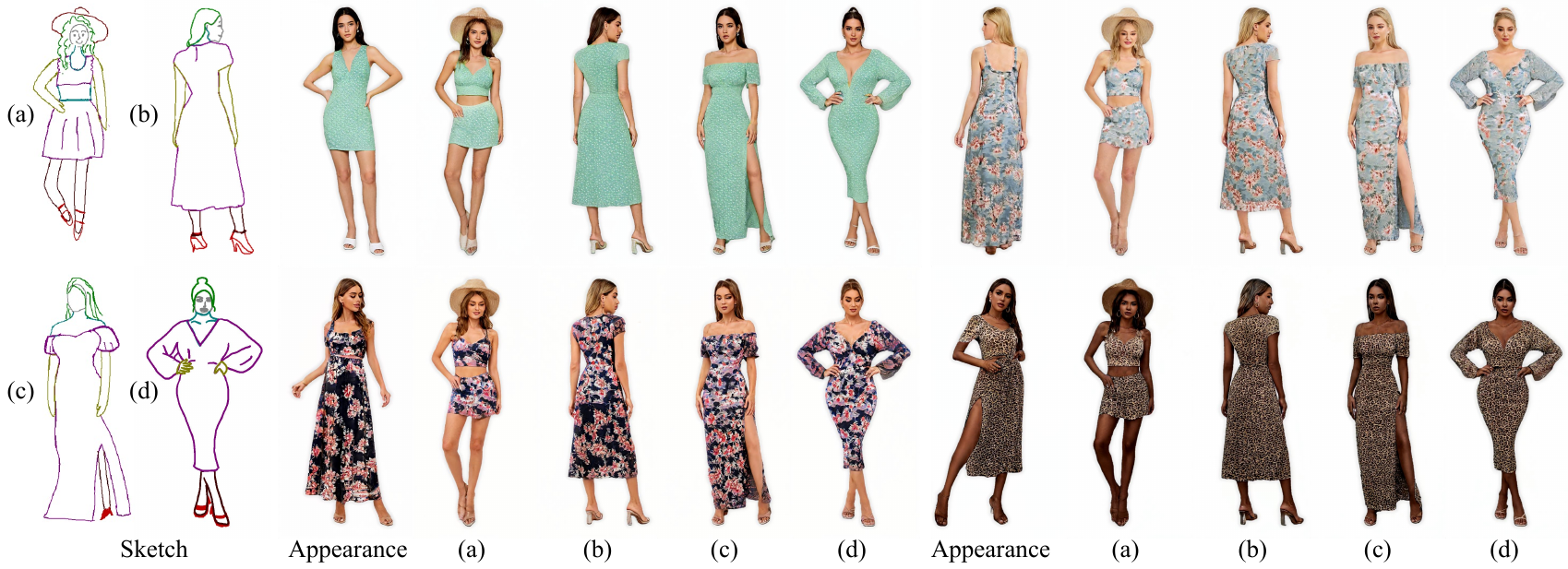}
  \caption{Our \sysName~ generates high-quality full-body images with respect to an input semantic sketch for geometry control and a reference image for appearance control. (a), (b), (c), and (d) correspond to four different sketch inputs.}
  \label{fig:teaser}
\end{figure*}

\IEEEPARstart{R}{ealistic} full-body human image
synthesis benefits various applications like fashion design \cite{dong2020fashion}, virtual try-on \cite{han2019clothflow,albahar2021pose,dong2022dressing,ren2022neural}, 2D avatar creation \cite{men2020controllable,fruhstuck2022insetgan}, and animations \cite{chan2019everybody, hong2022avatarclip}.
For such applications, high-fidelity generation and interactive control are both essential to generate specific images of interest. 
Although existing human image generation methods have produced impressive results, they are either unconditional \cite{fruhstuck2022insetgan, fu2022stylegan} or based on coarse conditions \cite{siarohin2018deformable, tang2021structure, albahar2021pose}. The unconditional methods produce high-fidelity and diverse images but lose controllability. The coarsely conditioned
methods achieve high-level control by coarse representations (\eg, pose \cite{tang2021structure,albahar2021pose}, text \cite{hong2022avatarclip, baldrati2023multimodal}, reference image \cite{han2019clothflow, han2018viton, choi2021viton, he2022style, dong2022dressing, kim2023reference}). However, these methods fail to explicitly and flexibly control detailed geometry (e.g., body contour, garment shape) 
\lzf{and appearance (e.g., skin color, garment texture) simultaneously. In terms of design, both novices and professionals with specific designs in mind often prefer a more subjective and specific control of results. }

\begin{figure}[htbp]
  \includegraphics[width=\linewidth]{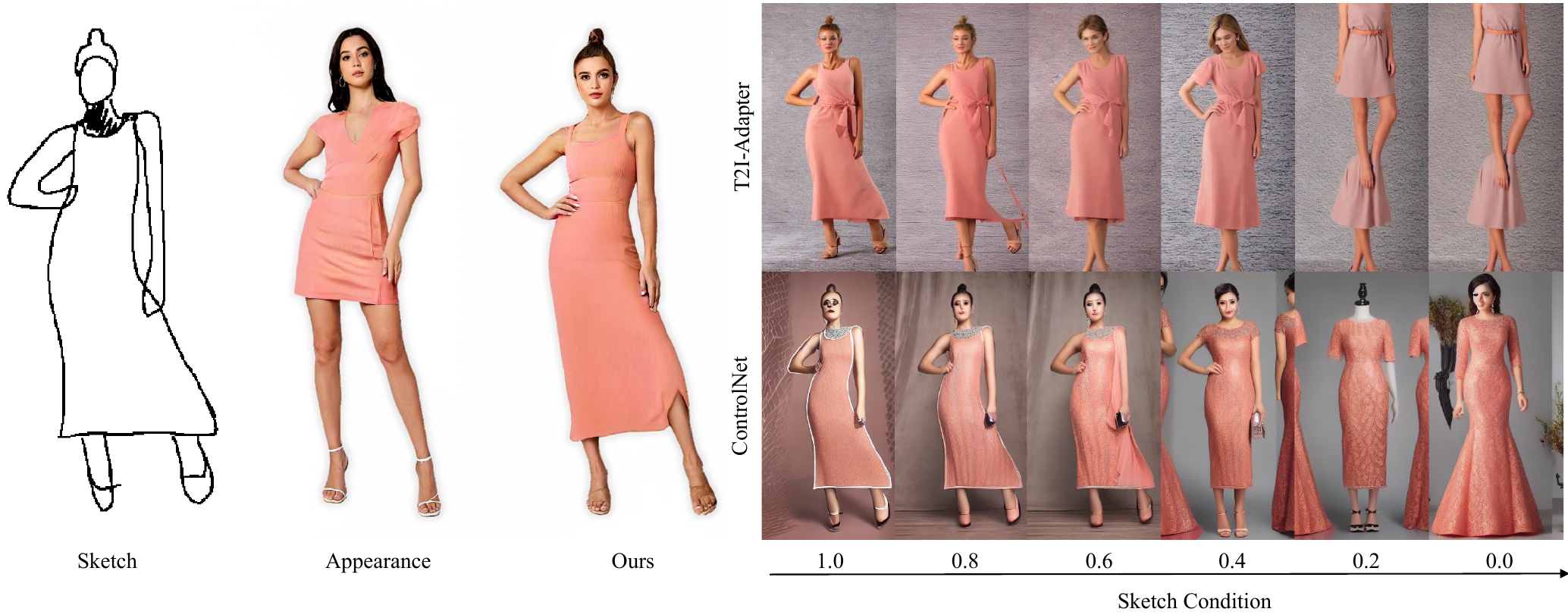}
  \caption{Qualitative comparisons with two diffusion-based methods with the decreasing 
  sketch conditioning weight. The text prompt for the two compared methods is "a woman wears a dress with salmon color".
  }
  \label{fig:sketch_con}
\end{figure}

Sketches are often used to explicitly depict desired geometry due to their simplicity, ease of modification, and ability to represent details. Sketch-based image generation techniques have been well explored in the human face domain \cite{chen2018sketchygan,li2020deepfacepencil,chenDeepFaceEditing2021}. However, sketch-based full-body human image generation is underexplored and more challenging than face generation since human bodies involve a larger variety of poses, garment shapes, and textures. These difficulties undoubtedly require a large amount of  data for training. Meanwhile, it is almost impossible to collect a real disentangled dataset including images with the same geometry but varying appearance or the same appearance with varying geometry for geometry and appearance control. Therefore, directly applying previous methods for human face generation to generate human bodies fails to produce high-fidelity results with diverse appearances (especially for subtle regions, \eg, face, shoes, glasses, and garment patterns) and preserve the reasonable human body structure. Benefiting from the prior knowledge in large text-to-image diffusion models \cite{rombach2022high, ramesh2022hierarchical}, ControlNet \cite{zhang2023adding} and T2I-Adapter \cite{mou2023t2i} leverage sketches to guide the multi-class object generation. 
However, with the increase of sketch abstraction, it is difficult to balance the photo-realism of the results and their faithfulness to
the input sketches (Figure \ref{fig:sketch_con})
and these methods cannot support consistent appearance control via appearance examples.

\begin{figure}
  \centering
  \includegraphics[width=\linewidth]{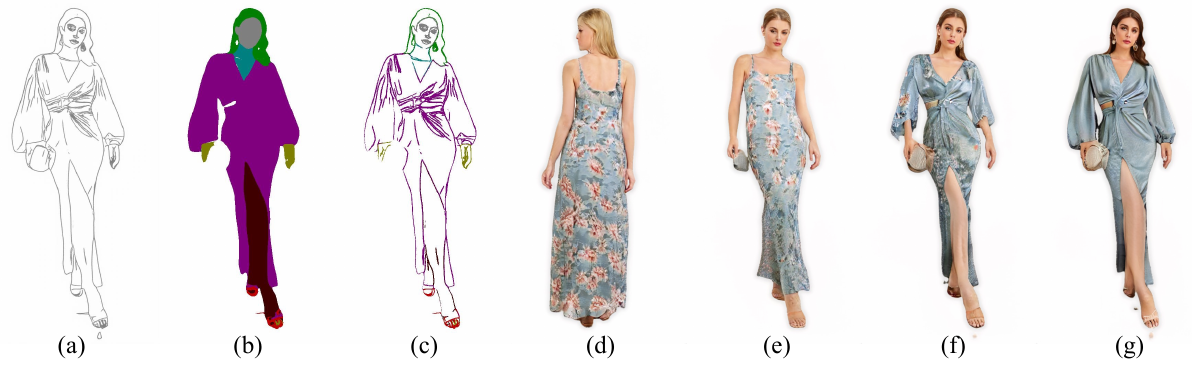}
  \caption{The examples of inputs and style-mixing results. (a) {a} 
  sketch input. (b) the corresponding parsing map. (c)
  the semantic sketch produced from (a) and (b). (d) {an input}
  appearance input. (e)-(g) results from mixing at layers 6, 8, and 10, respectively.}
  \label{fig:sample}
\end{figure}

To address these issues, we propose \emph{\sysName}, a novel deep generative framework for synthesizing a realistic human image from a semantic sketch (for flexible and detailed geometry control) and a reference appearance image (for appearance control), as shown in Figure \ref{fig:teaser}. Due to the lack of a disentangled full-body human dataset, we start from an unconditional generation model StyleGAN-Human \cite{fu2022stylegan}, which roughly disentangles the geometry and appearance in the latent space, as shown in Figure \ref{fig:sample}. To produce
photorealistic human images from sketches with different levels of abstraction, we embed the sketch in the latent space and then use StyleGAN-Human to generate results from such embeddings. To further promote the disentanglement of the StyleGAN-Human without a real dataset, we propose to employ the style mixing to automatically generate appearance-transferred and geometry-preserved training data (see Figure \ref{fig:sample} (e) and (g), respectively) that completely keep one constraint and associate it with the other constraint.

Specifically, we design a two-stage generation framework consisting of two main modules: \emph{Sketch Image Inversion} and \emph{Body Generator Tuning}. In Stage 1, we invert the input semantic sketch to a latent code via a sketch encoder. This encoder is trained with a large number of sampled images from StyleGAN-Human. To achieve accurate geometry inversion and avoid the influence of color and texture, we calculate sketch loss directly between the input sketches and the sketches extracted from the sampled images. Meanwhile, due to the sparsity of sketches, the sketch loss is not enough to help the encoder clearly distinguish each semantic part, especially for sketched humans with tight clothes. Therefore, we introduce a semantic loss. In Stage 2, we fine-tune the pretrained StyleGAN-Human with the appearance-transferred and geometry-preserved results, respectively. We leverage the style loss to learn the fabric feature from the appearance-transferred results and design a content loss using the geometry-preserved results to avoid geometric changes caused by the appearance-transferred results. We still leverage the above semantic loss to enhance geometry preservation. The large amount of synthesized data generated from the original StyleGAN-Human enables the full disentanglement between geometry and appearance.

Extensive experiments show that our method {achieves} 
flexible and disentangled control of geometry and appearance. Quantitative and qualitative comparisons prove that our \sysName~outperforms the related techniques for full-body image generation. We also demonstrate the robustness of our method against sketches of different styles by professionals and amateurs. 

\section{Related Work}
In this section, we review the existing techniques closely related to our method, including full-body human image synthesis and sketch-based generation and editing.

\subsection{Full-body Human Image Synthesis}
In recent years, 2D human generation has been widely investigated due to the rapid development of deep generative models.
The existing solutions can be roughly divided into unconditional and conditional methods. Unconditional methods aim to generate high-fidelity and diverse images from random noises. For example, from a
data-centric perspective, StyleGAN-Human \cite{fu2022stylegan} collects an SHHQ dataset to train a StyleGAN for the entire human body. InsetGAN \cite{fruhstuck2022insetgan} combines multiple pretrained GANs, each focusing on different parts (e.g., faces, shoes), which 
can be seamlessly merged into a full-body person image. Although the above methods generate realistic images, they lack explicit control of geometry or appearance. 

In contrast, conditional methods focus on controllable
generation via various conditions ({e.g.,} pose, reference images, text, etc.). 
\emph{Pose-conditioned} methods \cite{sarkar2021humangan, sarkar2021style, albahar2021pose} mainly depend on a canonical coordinate system of a 
3D human body (with UV parameterization) to directly establish the correspondence between pixels at each pose and then leverage a generator supervised with multi-view human datasets to refine coarse warped results. \emph{Virtual try-on} methods \cite{han2019clothflow, han2018viton, choi2021viton, he2022style, dong2022dressing} conditioned on reference images, aim to transfer garments in a reference person image to an input source person. 
They disentangle clothes from human identity at a segmentation generation stage and then warp clothes via a clothes deformation module to a 
target person. Compared with the pose-conditioned methods, virtual try-on methods further control the garment shapes and appearance via the reference images. But they are inflexible and inaccurate for simultaneous control of geometry and texture.
To facilitate intuitive control for layman users, Text2Human \cite{jiang2022text2human} utilizes an input text to constrain the shapes and textures of clothes. It first outputs a human parsing map from a given human pose and then synthesizes results guided by a text look-up code from a hierarchical texture-aware codebook. However, this approach limits the appearance to five predefined textures.

To achieve a more explicit and flexible geometry control, we adopt a semantic sketch as a 
geometry representation, which provides a concrete description and can be easily drawn and modified by novices and professionals. Different from the above virtual try-on methods, which use reference images to influence both geometry and appearance of generated images, our method uses a reference image to achieve appearance control, which is independent of geometry control by an input semantic sketch. 

\subsection{Sketch-based Image Generation and Editing}\label{sec:sketch}
Since it is easy for sketches to depict both global geometry (shape contours) and local geometry (e.g., wrinkles), sketch-based image generation and editing has been well explored. These methods have mainly focused on the human face domain. For example, 
DeepFaceDrawing \cite{chen2020deepfacedrawing}  and DeepFaceEditing \cite{chenDeepFaceEditing2021} leverage a local-to-global strategy: they first model separate features for each key face component and then recombine them together. Wu et al. \cite{wu2022deepportraitdrawing}
attempt to directly apply a local-to-global strategy to sketch-based human body generation. However, their approach still cannot achieve realistic results due to a larger variety of poses, shapes, and garments. In addition, their approach lacks any appearance control. Benefiting from the large text-to-image (T2I) models (\cite{rombach2022high, ramesh2022hierarchical}), \cite{zhang2023adding, mou2023t2i} learn different additional paths with geometry control inputs (e.g., sketch, pose, depth) to extract guidance features and then add them to the pretrained T2I models. \cite{baldrati2023multimodal, kim2023reference} directly concatenate the spatial controls with the original noises and finetune the pretrained T2I models in a self-supervised manner. However, \cite{zhang2023adding, baldrati2023multimodal} lack detailed control of appearance with text and \cite{mou2023t2i, kim2023reference} fail to faithfully transfer the appearance of reference images.

Exemplar-based methods \cite{zhou2021cocosnet, zhan2022marginal, liu2022dynast} are designed for controllable (geometry, appearance) image translation tasks and have been tested on multiple datasets (e.g., faces, indoor images). They use image retrieval to find appearance exemplars for training, thus alleviating the lack of a disentangled dataset, {including pairs with the same appearance but varying geometry.} 
They mainly learn a dense correspondence map between an input sketch and a reference image.
The learned correspondence might be inaccurate when large pose and shape differences exist between the sketch and reference image. Hence, such methods are unsuitable for generating human bodies with high complexity and diversity in the pose, body shape, and fashion elements. 

Recently, based on the layerwise representative of StyleGAN, embedding-based methods \cite{richardson2021encoding, su2022drawinginstyles} learn encoders to invert inputs to the corresponding latent space or latent features of StyleGAN. StyleGAN-Human is designed specifically for the full-body human domain. But due to incomplete disentanglement, it lacks control over clothing patterns with geometry unchanged. Since we would like to leverage the prior of StyleGAN, we also embed sketches to the latent space. To achieve accurate geometric embedding of complex human images, we train the encoder directly supervised by sketches and semantics rather than RGB images. Additionally, we synthesize 
appearance-transferred and geometry-preserved data to make StyleGAN-Human achieve 
full disentanglement.

\begin{figure*}[h]
  \centering
  \includegraphics[width=0.9\linewidth]{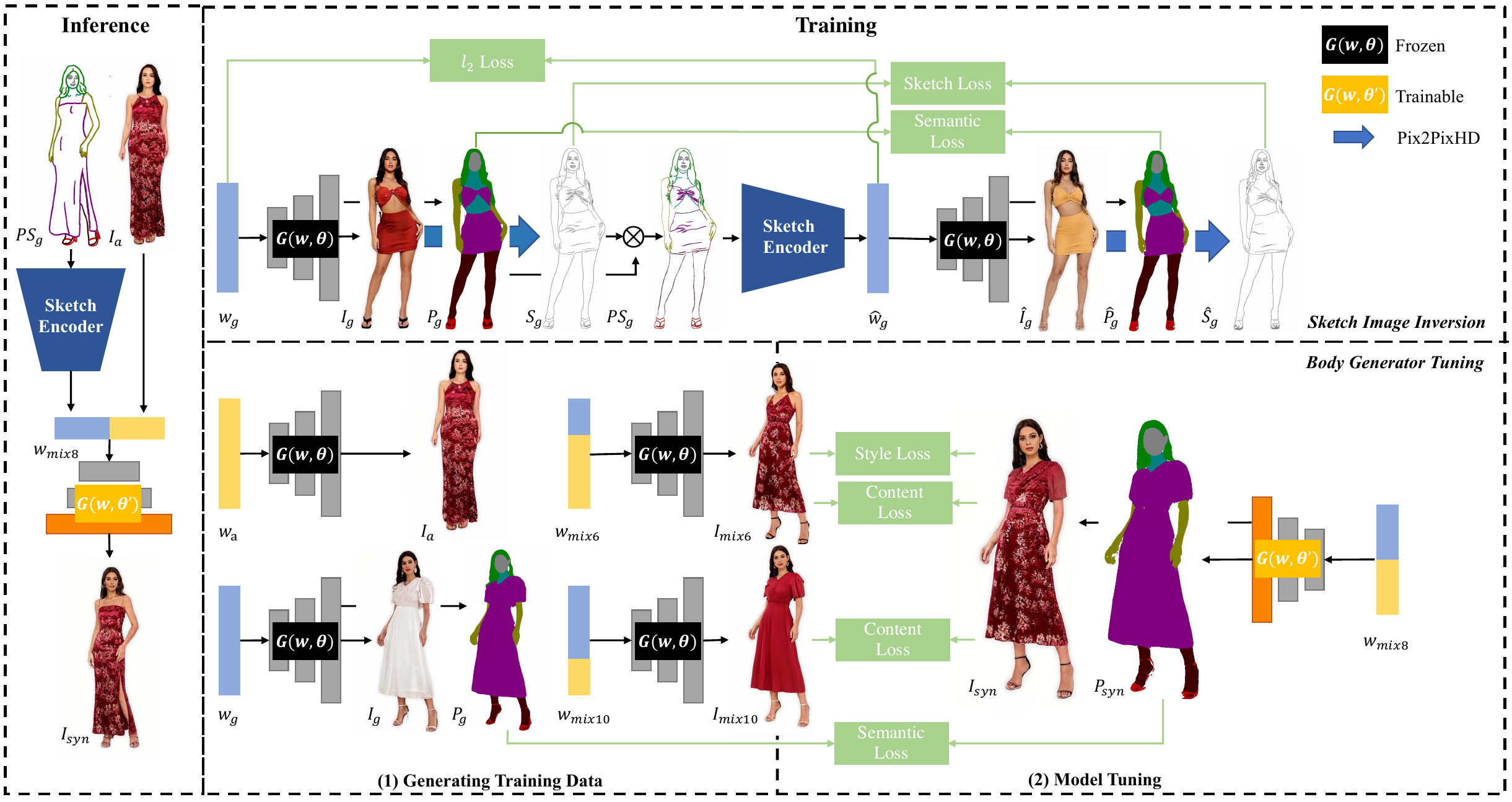}
\caption{An illustration about the training (right) and inference (left) pipelines of our method for full-body human image generation conditioned on a semantic sketch $PS_g$ and a reference image $I_a$. The training pipeline consists of two main modules: Sketch Image Inversion (right-top) and Body Generator Tuning (right-bottom). In the Sketch Image Inversion module, we first sample a latent code $w_g$ to generate the training triplet (semantic sketch $PS_g$, parsing map $P_g$, sketch $S_g$). Then, we use these data to train a sketch encoder. In the Body Generator Tuning module, given an appearance code $w_a$, we also sample a latent code $w_g$ to prepare the training appearance-transferred $I_{mix6}$ and geometry-preserved $I_{mix10}$ samples via style mixing at different layers. Then, we use them to fine-tune the generator $G(w; \theta^{'})$. During inference, the sketch encoder first embeds $PS_g$ into a latent code and mixes it with the appearance code derived from $I_a$ to form $w_{mix8}$. Given $w_{mix8}$, $G(w; \theta^{'})$ produces the final result $I_{syn}$.}
  \label{fig:architecture}
\end{figure*}

\section{Method}
For an input semantic sketch image $PS_g$ and an appearance image $I_a$, our \sysName~aims to synthesize a high-fidelity result $I_{syn}$ aligned with $PS_g$, while transferring the color and texture of face, skin, and clothing in $I_a$, as illustrated in Figure \ref{fig:architecture} (Left). The semantic labels in our current implementation include nine categories: hat, hair, hand, glasses, garment, torso-skin, face, foot, and shoe. Users might input semantic sketches stroke by stroke, with each stroke associated with a specific semantic label (see Supplementary Materials Section IV-A).

As illustrated in Figure \ref{fig:architecture} (Right), our system consists of two main modules, namely, Sketch Image Inversion (Section \ref{sec:sketch}) and Body Generator Tuning (Section \ref{sec:body}). To produce photo-realistic full-body human images from sketches with different styles, the first module focuses on inverting an input semantic sketch $PS_g$ to a geometry latent code $\hat{w}_g$, located in the $W+$ space. Compared with the $W$ space, the $W+$ space with 18 style vectors can represent more accurate details of the input. Due to the large variability of real human-body images, StyleGAN-Human fails to restore complex fabric patterns accurately when it roughly maintains geometry input (Figure \ref{fig:sample} (f)). Note that a na\"{i}ve approach by fine-tuning on the appearance image cannot obtain satisfactory results due to the large variance in the pose and shape (Section \ref{sec:ab2}). Therefore, we propose to use a novel training strategy to fine-tune the generator in the second module.

\subsection{Sketch Image Inversion}\label{sec:sketch}
At this stage, based on the pre-trained StyleGAN-Human $G(w; \theta)$, we train a sketch encoder responsible for embedding the geometry information of $PS_g$ as comprehensively as possible. {Although an input sketch could be converted to multiple latent codes with the same geometry and different appearance, the geometric parts of those multi-latent codes are the same. Thus, the sketch encoder only focuses on accurate geometry one-to-one mapping (sketch to the geometric part). Specifically, it embeds an input semantic sketch $PS_g$ to a latent code $\hat{w}_g \in {18 \times 512}$, as shown in Figure \ref{fig:architecture}.} 
During training, $PS_g$ comes from a combination of an automatically generated sketch $S_g$ and a parsing map $P_g$. During inference, users interactively provide semantic strokes $PS_g$, as shown in the accompanying video. Since the feature pyramid structure of Encoder4Editing (e4e) encoder \cite{tov2021designing} makes it possible to  encode the input details, we adopt its structure for our encoder.
StyleGAN-Human cannot fully capture the distribution of complex real full-body images because such images might be out-of-distribution for it. If trained with those real images, the encoder would introduce unreasonable structures into $\hat{w}_g$. Then such artifacts would be displayed after the style-mixing (Figure \ref{fig:cm_sketch}). To avoid this
issue, we train the encoder with paired data $\{(w_g, I_g)\}$ sampled from the latent space of StyleGAN-Human.

\subsubsection{Training}
To provide the supervision directly on the sketches, we retrain Pix2PixHD \cite{wang2018high} for the image-to-sketch generation task with the paired data $\{(I, S)\}$. The images $\{I\}$ are from the SHHQ dataset \cite{fu2022stylegan}, and {the corresponding sketches $\{S\}$ are extracted by the Sobel filter and the sketch simplification method \cite{simo2016learning}.} The network structure and training process are unchanged. Then, Pix2PixHD generates sketches $\{S\}$ from input images $\{I\}$, as illustrated by the blue arrow in Figure \ref{fig:architecture}. 
Inspired by \cite{li2021semantic}, we add a semantic branch StyleGAN-Human to facilitate semantic label extraction. The detailed architecture is shown in Supplementary Materials Section I.
Benefiting from the underlying semantic information in StyleGAN-Human, it is easy to learn such a semantic branch with sampled pairs $\{(w, I)\}$. Here, we provide the semantic label $P$ from $I$ using an off-the-shelf method \cite{gong2019graphonomy}. After training, the generator $G(w; \theta)$ directly produces $(I, P)$ for any input latent code $w$.

For the specific training procedure of the encoder, we randomly sample a latent code $w_g$ and feed it into $G(w; \theta)$ to generate a synthetic full-body image $I_g$ and a corresponding parsing map $P_g$. Meanwhile, 
Pix2PixHD produces the corresponding sketch image $S_g$ from the input image $I_g$. As shown in Figure \ref{fig:architecture}, we transfer the semantics from the parsing map $P_g$ to the sketch image $S_g$ via pixel-wise multiplying them together to get a semantic sketch $PS_g$, which is then projected to $\hat{w}_g$ via our sketch encoder. Finally, inputting the whole $\hat{w}_g$ to $G(w; \theta)$, the generator outputs a synthesized image $\hat{I}_{g}$ and a parsing map $\hat{P}_{g}$. We still use Pix2PixHD to produce a sketch image $\hat{S}_{g}$.

\subsubsection{Objective Function}
Based on the frozen generator $G(w; \theta)$, the goal of our sketch encoder is to reconstruct the geometry information in 
$S_g$ with a low error. To achieve this, we formulate the objective function from three aspects: (1) distribution consistency, (2) geometric accuracy, and (3) semantic consistency.

\textbf{Distribution Consistency.} To encourage the geometry codes follow the distribution of the $W+$ space, we first adopt the latent adversarial loss $L_{adv_w}$ proposed in e4e \cite{tov2021designing}. We also use the $l_{2}$ loss $L_{l_2}$, which is calculated between the inverted code $\hat{w}_g$ and the ground-truth code $w_g$. This simple way
makes the encoder easy to project the inputs to the correct distribution, thus alleviating unreasonable embeddings.

\textbf{Geometry Accuracy.}
Sketching inputs only contain geometric information. Our sketch encoder is not too concerned with the code related to the appearance for the high-resolution ($64^2-1024^2$) layers. Compared with calculating losses on RGB images and thus indirectly constraining geometry, it is more efficient to directly supervise on the sketches. Benefiting from the image-to-sketch model Pix2PixHD, we can get $\hat{S}_g$ from the generated image $\hat{I}_g$. Liu et al. \cite{liu2022deepfacevideoediting} found that LPIPS \cite{johnson2016perceptual} is effective for sketches due to its sensitivity for edges. Hence, we introduce the sketch loss $L_{lpips}$, which aligns the features from VGG \cite{simonyan2014very}, given the respective inputs of a sampled sketch $S_g$ and the corresponding generated sketch $\hat{S}_g$.

\textbf{Semantic Consistency.}
Compared with RGB images, sketches are sparse and lack color information. Such an issue makes the encoder trained with sketches themselves easier to confuse various semantic parts (Section \ref{sec:ab1}), especially when the body and clothes share similar contours (e.g., due to wearing tight clothes). Hence, we incorporate a semantic loss to force the encoder to learn the semantic correspondence between the input sketch and the latent code. Specifically, the semantic loss consists of the pixel-wise cross-entropy loss $L_{ce}$ and the dice loss $L_{dice}$, which are generally used in the semantic segmentation task \cite{isensee2018nnu}. 
We calculate those two terms between the parsing maps $P_{g}$ and $\hat{P}_{g}$ derived from the input sketch and the reconstructed result, respectively.

By combining the above loss terms, we define the final objective function as follows:
\lzf{\begin{equation}
\label{eq:encoder loss}
\begin{split}
L_{encoder} = &{\lambda_1}{L_{adv_w}(\hat{w}_g)} + {\lambda_2}{L_{l_2}(w_g, \hat{w}_g)} + {\lambda_3}{L_{lpips}(S_g, \hat{S}_g)} \\
&+ {\lambda_4}{L_{ce}(P_{g}, \hat{P}_g)} + {\lambda_5}{L_{dice}(P_{g}, \hat{P}_g)}. 
\end{split}
\end{equation}
In our experiments, we set ${\lambda_1=0.1}$, ${\lambda_2=10}$, ${\lambda_{3}=2}$, ${\lambda_{4}=1}$, and ${\lambda_{5}=1}$.}

\subsection{Body Generator Tuning}\label{sec:body}
Our body generator follows a standard structure of StyleGAN-Human. The model has two conditions (i.e., the inputs): a geometry code $w_g$ to provide geometry constraints and an appearance code $w_a$ to provide appearance constraints.
StyleGAN is composed of multiple style blocks, and each block is responsible for generating a specific level of detail in the image. These style blocks are associated with latent codes that control the different attributes. In style-mixing, different blocks can be assigned distinct codes. 
This allows for combining 
styles from multiple codes, resulting in diverse generated images. 
StyleGAN-Human observes that the high-layer styles (9-18) control the clothing color, middle-layer styles (5-8) control the clothing type and human face identity (geometry and appearance), and low-layer styles (1-4) control the pose. Hence, we follow the mixing rules between the high layers and the rest layers. The model $G(w; \theta^{'})$ combines the geometry and appearance information by mixing these two codes at layer 8, denoted as $w_{mix8}$, to synthesize the result $I_{syn}$. As shown in Figure \ref{fig:sample} (f), applying the mixed code $w_{mix8}$ to the original generator $G(w; \theta)$ produces an image that slightly changes the geometry from $S_g$ and only exhibits the color but not texture from $I_a$. To faithfully align the geometry and restore the texture, for each appearance image, we first generate a corresponding training dataset (Section \ref{sec:generate_data}) and then fine-tune the original generator $G(w; \theta)$ to a new body generator $G(w; \theta')$, which achieves a more complete disentanglement (Section \ref{sec:model_tune}). The key idea is to leverage the style mixing of an unconditional GAN to automatically synthesize the training data.

\subsubsection{Generating Training Data}\label{sec:generate_data}
Intuitively, fine-tuning with a single appearance image $I_a$ using a style loss \cite{johnson2016perceptual} can help the generator learn the appearance. However, we found several issues with such a na\"{i}ve approach (Section \ref{sec:ab2}). First, the model trained with the style loss tends to learn the mean textures and sometimes blends the textures from different semantic parts. Second, the appearance of clothes is pose-dependent, but transferring the style from this single appearance image $I_a$ just copies patterns under the original pose. Meanwhile, it is challenging to build a dataset with different geometry for the same appearance.

We observe that by copying the geometry code for layers (1-6) and the appearance code for the rest layers, the results restore the reference appearance well with the pose indicated in the input geometry code but does not faithfully respect the detailed geometry information
(e.g., garment change in Figure \ref{fig:sample} (e)). We call such results the \emph{appearance-transferred} results. On the other hand, when inputting the geometry code for layers (1-10), the results preserve the global and local geometry well but only gain the color without any fabric pattern from the appearance code (Figure \ref{fig:sample} (g)). We thus refer to such results as the \emph{geometry-preserved} results. Although the two kinds of style-mixing strategies we find above
fully maintain only geometry or appearance information, they can help the generator achieve a comprehensive disentanglement of geometry and appearance.

\begin{figure}[t]
  \centering
  \includegraphics[width=\linewidth]{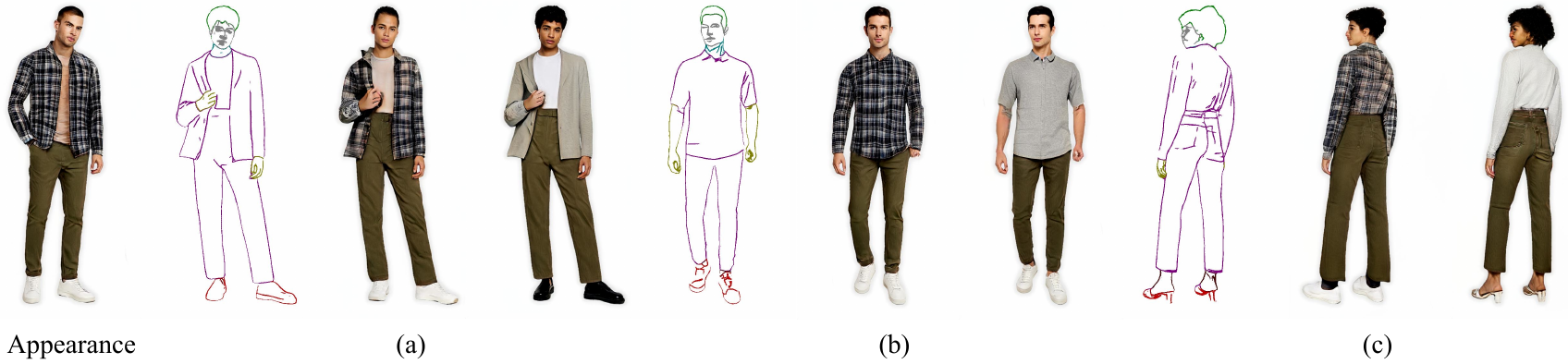}
  \caption{Three examples of prepared data. Each example in (a)-(c) shows an input semantic sketch and its corresponding appearance-transferred and geometry-preserved results.}
  \label{fig:data}
\end{figure}

Hence, we first sample a large number of latent codes $\{w_g\}$ from the latent space of the generator $G(w; \theta)$ to represent different geometry. Then, for each appearance input $I_a$, we generate the synthesized appearance-transferred $\{I_{mix6}\}$ and geometry-preserved $\{I_{mix10}\}$ pairs by feeding the pre-trained generator $G(w; \theta)$ with the mixed codes between the appearance code $w_a$ and geometry codes $\{w_g\}$ at layers 6 and 10 separately (Figure \ref{fig:architecture}). The representative training examples are shown in Figure \ref{fig:data}.

\subsubsection{Model Tuning}\label{sec:model_tune}

The goal of our body generator is to restore the whole appearance information (color and texture) on the premise of preserving the geometry information defined in the latent code $\hat{w}_g$ obtained in Section \ref{sec:sketch}. We freeze the affine transformation layers since we need to utilize the original latent space. Additionally, Alaluf et al. \cite{alaluf2022hyperstyle} find that altering the \emph{toRGB} layers harms the editing capabilities of GAN. Therefore, we only tune the \emph{non-toRGB} convolution layers. 

For appearance learning, as illustrated in Figure \ref{fig:architecture}, we first incorporate the style loss $L_{style}$ \cite{johnson2016perceptual} between each semantic region $i$ of the generated image $I_{syn}$ and target image $I_{mix{6}}$ via computing the Gram matrix for the features extracted by VGG \cite{simonyan2014very}. Those regions are extracted from the corresponding parsing maps $P_{syn}$ and $P_{mix6}$. Note that the generated images $\{I_{syn}\}$ and $\{I_{mix6}\}$ are not entirely aligned in terms of geometry, especially for the clothing shapes. Hence, we calculate a content loss $L_{content_6}$ \cite{johnson2016perceptual} which involves spatial features within each semantic union region via a union mask $M$. $M_i$ refers to the overlap regions between semantic label $i$ of $P_{mix6}$ and $P_{syn}$. The extraction of the $M$ is illustrated in Supplemental Materials Section II.

However, only using the above loss harms the preservation of the geometry information since respecting the image $I_{mix6}$ more or less changes the geometry defined by the input sketch. It is important for our body generator $G(w; \theta^{'})$ to promote the ability to restore the geometric requirements. Hence, we propose geometric constraints from three aspects. First, since the different layers of StyleGAN-Human control different visual attributes, we update the (9-18) convolution layers related to the appearance code, leaving the (1-8) convolution layers related to the geometry unchanged. Such a constraint can not only preserve the geometry prior but also reduce the training parameters, thus saving training time.

Second, benefiting from the semantic branch mentioned in
Section \ref{sec:sketch}, we get the paired parsing maps with the synthesized results simultaneously. When inputting the whole $w_g$, the generator provides the corresponding parsing map $P_g$. Similarly, we can get $P_{syn}$ with the input of $w_{mix8}$. Then, we only measure the dice loss $L_{dice}$ \cite{isensee2018nnu} between them, since we find that the additional \emph{ce} loss here is harmful to the visual quality.

Third, semantic supervision only focuses on maintaining the global geometry but cannot preserve local features (e.g., wrinkles), thus decreasing the generation quality. Therefore, we calculate the high-level content loss $L_{content_{10}}$ to encourage the output image $I_{syn}$ to be perceptually consistent with the geometry-preserved image $I_{mix{10}}$ in terms of the image content and spatial structure.

The final loss objective is defined as follows:
\begin{equation}
\label{eq:decoder loss}
\begin{split}
L_{decoder} = &{\lambda_6}{\sum_{i=1}^n L_{style}( P_{mix6,i} \cdot I_{mix6}, P_{syn,i} \cdot I_{syn})} \\
&+{\lambda_7}{\sum_{i=1}^n L_{content_6}}(M_i \cdot I_{mix6}, M_i \cdot I_{syn})\\
&+ {\lambda_8}{L_{content_{10}}(I_{mix{10}}, I_{syn})} + {\lambda_9}{L_{dice}(P_g, P_{syn})}, 
\end{split}
\end{equation}
where ${\lambda_6=1000}$, ${\lambda_7=0.6}$, ${\lambda_8=1}$, and ${\lambda_9=30}$. $n$ is the number of semantic labels.

\section{Experiment}
\begin{figure*}[h]
  \centering
  \includegraphics[width=\linewidth]{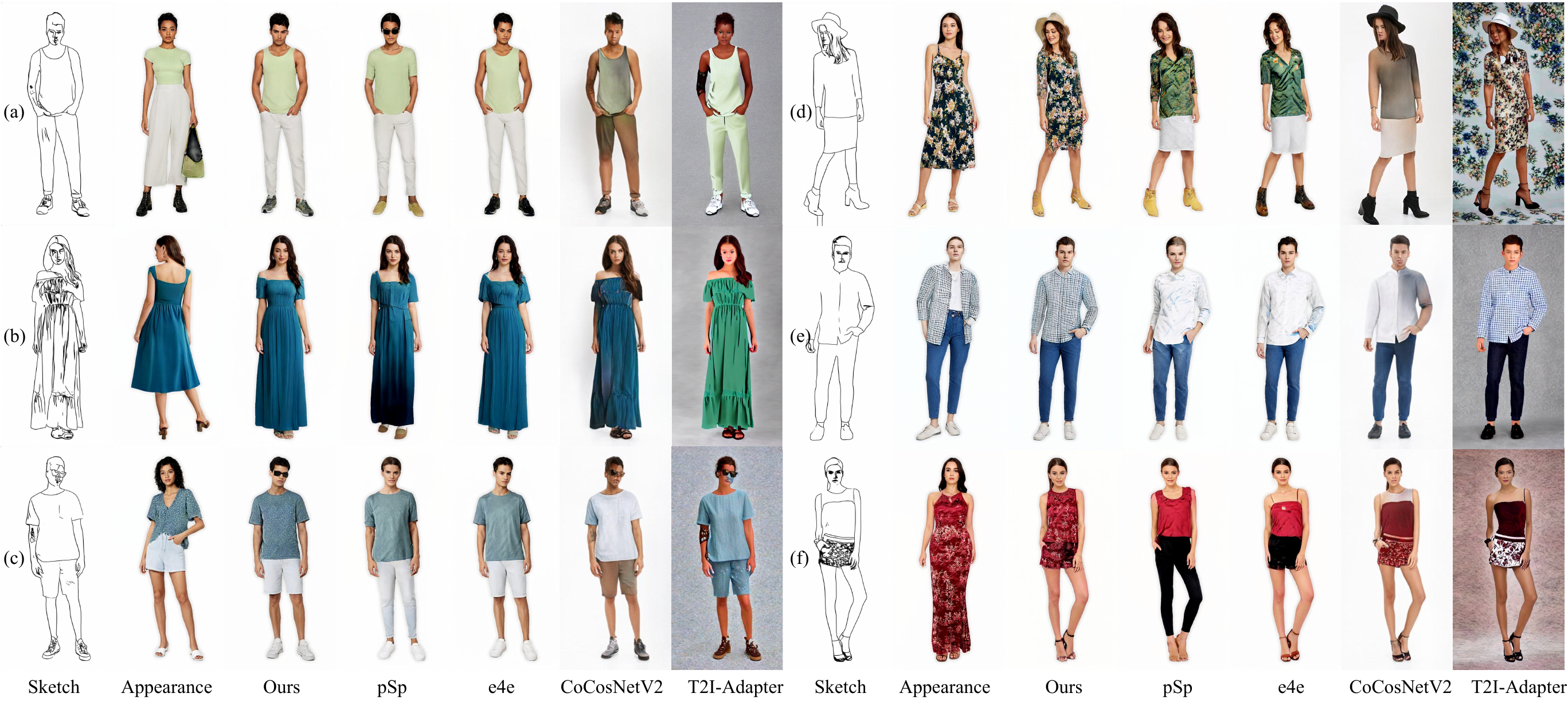}
  \caption{Qualitative comparisons between our method and four related sketch-based methods. Our method shows the best geometry and appearance transfer results. The sketch images are extracted from the DeepFashion dataset. The appearance images sampled from the StyleGAN-Human include pure color and texture images. Pure color images denote garments containing one or more colors without fabric patterns (a)-(b), while texture images include both (c)-(f).}
  \label{fig:cm_sketch}
\end{figure*}

\begin{figure*}[h]
  \centering
  \includegraphics[width=0.9\linewidth]{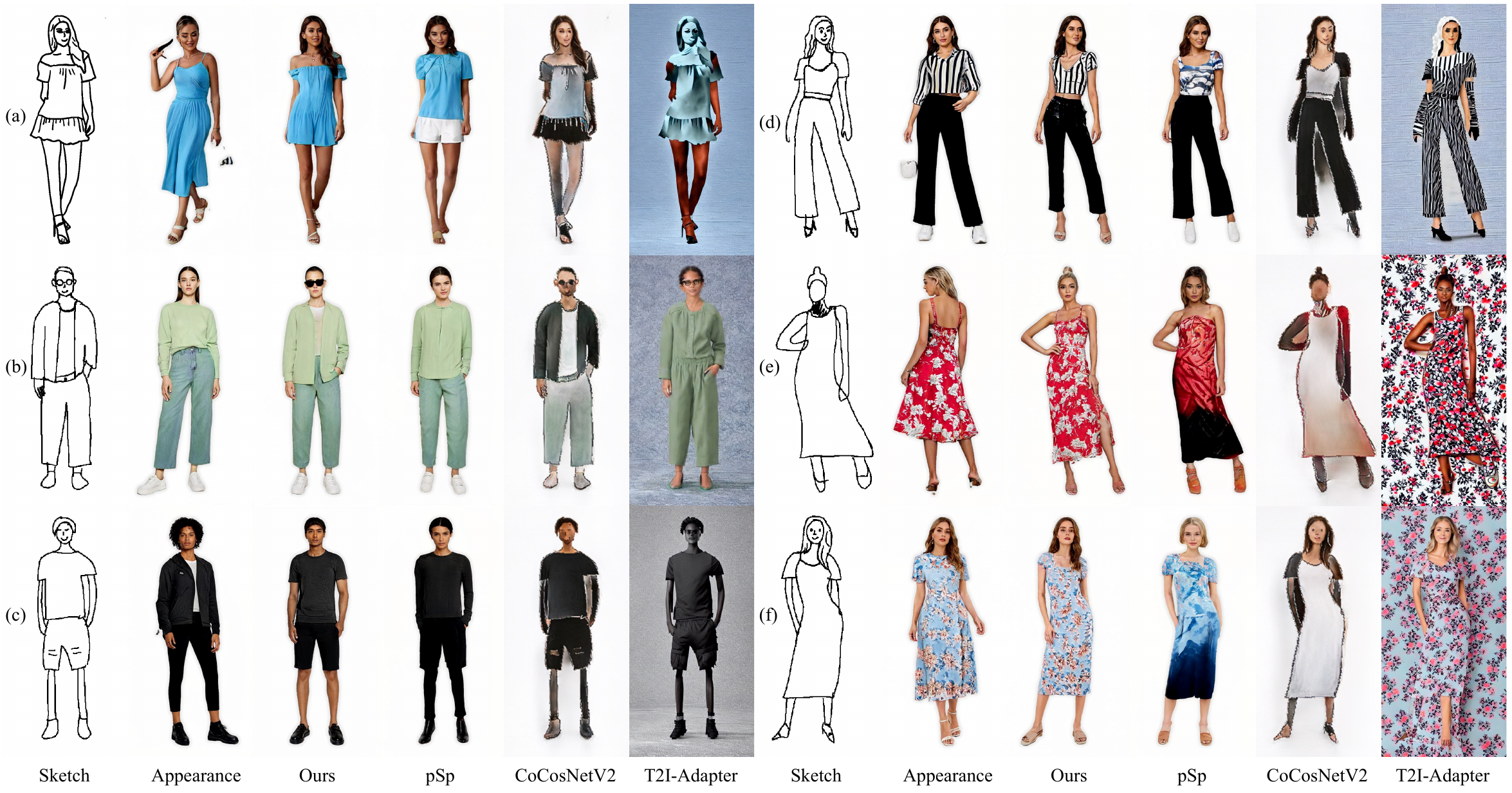}
  \caption{Qualitative comparisons between our method and three 
  related methods (without e4e since it takes as input RGB images). Our method shows high-fidelity results with the best geometry and appearance consistency. The sketch images are collected from users. }
  \label{fig:cm_user}
\end{figure*}

In this section, we have done extensive experiments from five aspects, namely, baseline comparison (Section \ref{sec:cm}), face body montage with InsetGAN (Section \ref{sec:insetgan}), real appearnce image (Section \ref{sec:real}), user study (Section \ref{sec:user}) and ablation study (Section \ref{sec:ab}). For the quantitative comparison, we leverage the symmetric \emph{Chamfer Distance} (CD) \cite{wang2022rewriting} and \emph{Mean Intersection over Union} (MIoU) to evaluate the geometry alignment with the sketch input. The visual quality is measured by the $Fr\acute{e}chet \ Inception \ Distance$ (FID) \cite{heusel2017gans} and Inception Score (IS) \cite{salimans2016improved}. For the user study, each participant was asked to consider four aspects, including the accuracy of geometry preservation (GP), appearance transfer (AT), visual quality (VQ), and user preference (UP), and then choose the best one for each aspect. The details about system implementation and evaluation metrics can be found in Supplementary Materials Section II.

\begin{table}[t]
\begin{center}
 \begin{tabular}{ccccc}
 \hline
	\multicolumn{1}{c}{\textbf{Method}}
	& \multicolumn{1}{c}{\textbf{CD $\downarrow$}} 
        & \multicolumn{1}{c}{\textbf{MIoU $\uparrow$}} 
        & \multicolumn{1}{c}{\textbf{FID $\downarrow$}}
	& \multicolumn{1}{c}{\textbf{IS $\uparrow$}} \\
	\hline
	pSp  & 4.66 & 0.72 & 41.38 & 2.69\\
        e4e   & 5.06 & 0.76 & 22.64 & 2.73\\
        \lzf{CocosNetV2 } & \textbf{1.07} & \textbf{0.83} & 39.09 & 2.71\\
        \lzf{T2I-Adapter}  & 3.22 & 0.74 & 56.73 & \textbf{3.74}\\
        \hline
        Ours  & 4.95 & 0.80 & \textbf{22.04} & 3.05\\
        \hline
\end{tabular}
\end{center}
\caption{Quantitative comparison between  our \sysName~and four related methods on the DeepFashion dataset \cite{liu2016deepfashion}.}
\label{tab:qua_cm}
\vspace{-3em}
\end{table}

\subsection{Baseline Comparison}\label{sec:cm}
\subsubsection{Qualitative Comparison}\label{sec:qual}
To the best of our knowledge, no prior work generates full-body human images conditioned on geometry and appearance images. Hence, we first re-purpose the related image generation methods with the above two inputs, including pSp \cite{richardson2021encoding}, e4e \cite{tov2021designing}, CoCosNetv2 \cite{zhou2021cocosnet}, and T2I-Adapter \cite{mou2023t2i} for our controllable full-body generation task. The evaluation details of the above methods and text prompts for T2I-Adapter are in Supplementary Materials Section III-A.

\textbf{Geometry Quality.} Figures \ref{fig:cm_sketch} and \ref{fig:cm_user} show the results from fine and coarse sketches separately.
pSp can basically capture the global geometry of sketch images varied in abstraction. Still, this approach tends to add or lose semantic parts, resulting in salient artifacts (e.g., glasses in Figure \ref{fig:cm_sketch} (a) and hat in Figure \ref{fig:cm_sketch} (d)). Additionally, the local geometry is rough since several strokes are ignored (see the pSp results in Figure \ref{fig:cm_user} (c) and (f)). In addition, the geometry code from pSp sometimes affects subsequent color expression (Figure \ref{fig:cm_sketch} (b) and Figure \ref{fig:cm_user} (e)). 
This is mainly because the predicted embeddings might be out-of-distribution. The e4e can effectively infer latent codes that belong to a latent space distribution because of the latent discriminator. So the visual quality of e4e's results is relatively high. But their results might fail to capture the accurate postures (see the e4e results in Figure \ref{fig:cm_sketch} (a)) and local details (see Figure \ref{fig:cm_sketch} (b) and (d)) possibly due to the co-learning of the geometry and appearance. CocosNetV2 completely aligns the geometry of results with the sketches. Thus, it fails to synthesize realistic full-body images from coarse sketches with inaccurate human proportions and shapes (see the CocosNetV2 results in Figure \ref{fig:cm_user}). T2I-Adapter generates satisfactory results even with rough input sketches since it leverages an adjustable weight to control the geometry consistency. But with the almost full geometric alignment, they produce unrealistic faces (see the T2I-Adapter results in Figure \ref{fig:cm_user} (a) and (d))  and fail to restore the real body proportion (see Figure \ref{fig:cm_user} (c) and (e)), which are not included in the input. Benefiting from the large sampled data and the explicit supervision of geometry and semantics in our network, our results better preserve the global geometry (poses, clothing types) and synthesize each semantic element more faithfully. 

\textbf{Appearance Quality.} As for the appearance transfer, pSp, e4e, and CocosNetV2 can only transfer the primary colors of the appearance inputs. The reason for pSp and e4e is that the high layers of StyleGAN-Human can only control the colors. CocosNetV2 trained with the DeepFashion dataset fails to build accurate dense correspondence when the sketch input does not depict the texture boundaries. T2I-Adapter regards the appearance input as a style, so it cannot accurately restore appearance details, and their results tend to be non-photorealistic (see the T2I-Adapter results in Figure \ref{fig:cm_user} (a) and (d)). As embedded in the latent space of StyleGAN-Human, our method can generate high-fidelity results from different degrees of sketches. Meanwhile, thanks to altering the generator with the roughly appearance-transferred data, our method can best transfer color and texture simultaneously from the reference images.

\begin{figure*}[h]
  \centering
  \includegraphics[width=\linewidth]{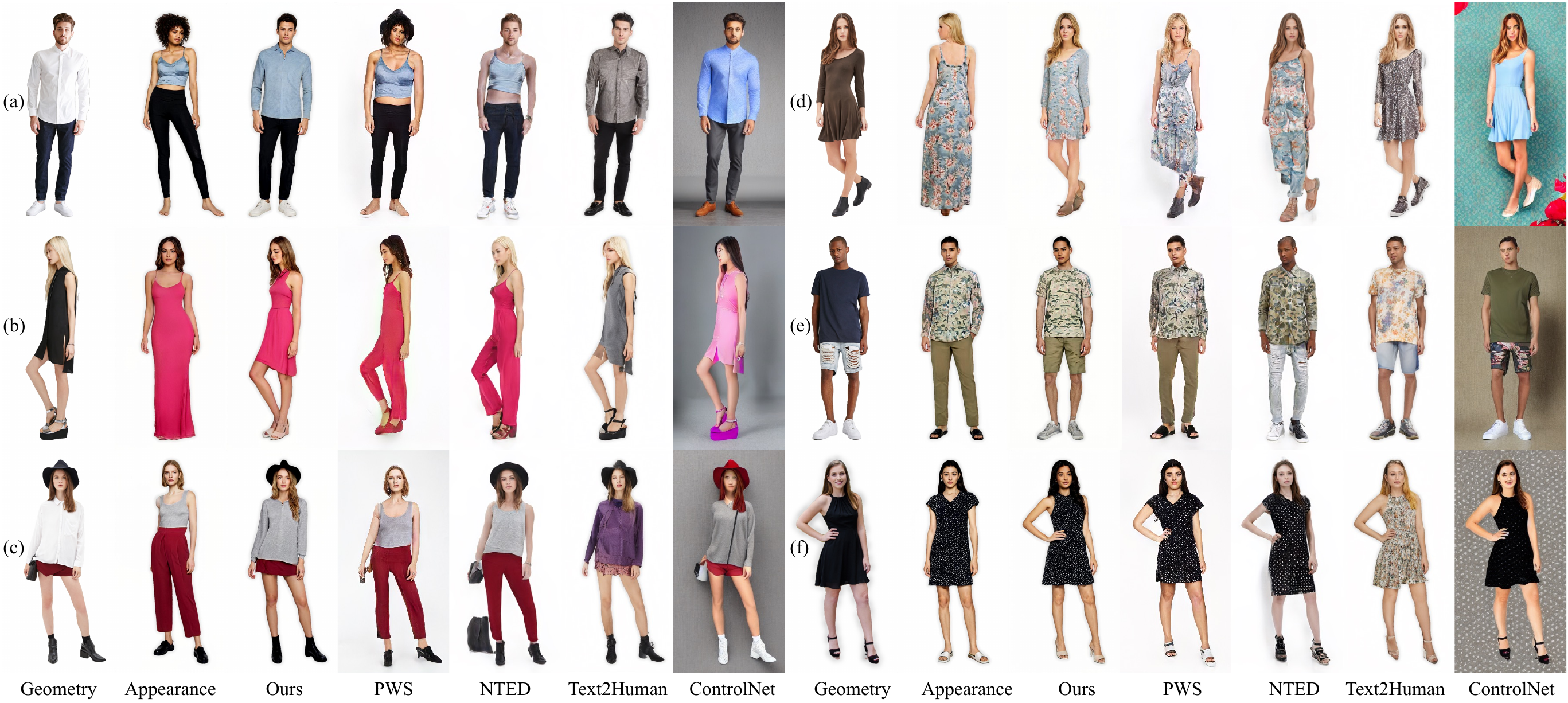}
  \caption{Qualitative comparisons of our method with four human image generation methods. Our method demonstrates the most flexible and detailed control in both geometry and appearance.}
  \label{fig:cm_body}
\end{figure*}

\subsubsection{Comparison with Human Image Generation Methods}\label{sec:cm_body}
To prove the flexibility and detailed control of our inputs, we also compare our method with full-body image generation methods with different inputs, including PWS \cite{albahar2021pose}, NTED \cite{ren2022neural}, Text2Human \cite{jiang2022text2human}, and ControlNet \cite{zhang2023adding}. Although the modalities of their inputs are different, we derive the inputs required by different methods from the same RGB images. The details for input extraction and text prompts for ControlNet are in Supplementary Materials Section III-B.

As shown in Figure \ref{fig:cm_body}, our method achieves the highest fidelity. PWS can only change the human pose in a given appearance image, while NTED aims to transfer the garments in the appearance image to the human in the geometry image. Besides a pose and specific cloth, our method can also change the hairstyle, garment shape, etc. 
{For the virtual try-on effect similar to NTED, it is vital to retain the face completely (i.e., both geometry and appearance) of the geometry input image, while our method is designed to retain only the geometry of the geometry image (Figure \ref{fig:cm_body} Column \emph{Ours}) since we assume the appearance is from the reference image. Our method could potentially achieve this application via drawing a similar clothing type with the reference image and directly swapping the face of the geometry image with our result (e.g., by using InsetGAN) (Figure \ref{fig:insetgan} (a)). 
Text2Human and ControlNet take parsing maps and canny maps as input, respectively. They achieve satisfactory geometry preservation. However, the appearance of these two methods is mainly controlled by text. Text2human generates the results with a completely irrelevant appearance since it supports only five types of appearance text (e.g., stripe, plaid). Based on a large text-to-image model, ControlNet produces more specific results from sufficient text prompts. But compared with a given appearance image, a text prompt is still too coarse to describe detailed colors and textures.

\subsubsection{Quantitative Comparison}\label{sec:qua_cm}
Table \ref{tab:qua_cm} shows the quantitative comparison results. Since the results of CocosNetV2 are strictly aligned with the input sketches (Figure \ref{fig:cm_sketch} and \ref{fig:cm_user}), it achieves the highest CD and mIoU. Here we set the weight of T2I-Adapter to 1, so it produces the comparable CD. However, their method might ignore several strokes located in the semantic boundary (e.g., Figure \ref{fig:cm_sketch} (d), (f) and Figure \ref{fig:cm_user} (e), (f)), resulting in a low MIoU. Since T2I-Adapter is based on a
large text-to-image model, its results tend to exhibit more diversity according to reference images, achieving the highest IS. But its results fail to preserve the color and texture in the target references.
Additionally, CocosNetV2 and T2I-Adapter lack realism, as reflected in their high FID scores. The realism tends to be worse when the input sketches get rougher (Figure \ref{fig:cm_user}). 
For a fair comparison between the embedding-based methods, we use the same generator, i.e., StyleGAN-Human, for testing. For FID, our method has achieved a similar result to e4e, proving that our encoder trained with the sampled data does not affect the fidelity of the generator. The CD of pSp is lower than us, but this is inconsistent with our observations of more serious artifacts with their results, as shown in Figure \ref{fig:cm_sketch}. Meanwhile, the highest FID of pSp also reflects the lowest quality of synthesized images. 
Our method significantly outperforms the other four methods in terms of FID with comparable IS and mIoU. This indicates that our results are not only the most similar to the sketch inputs but also of the highest image quality. This is consistent with our findings based on the qualitative comparisons.

\subsection{Face Body Montage with InsetGAN}\label{sec:insetgan}
Since our method combines the geometry and appearance information from two inputs, the face identity (involving both geometry and appearance) of the reference image 
might be lost. However, several tasks (e.g., virtual try-on, human image editing) require preserving human face identity. Benefiting from InsetGAN \cite{fruhstuck2022insetgan}, which introduces a multi-GAN optimization method, 
{our pipeline can be easily combined with the state-of-art face model \cite{karras2020analyzing} to achieve face identity preservation.} To keep the identity of the reference image or geometry image (used to extract a sketch input), we first invert the corresponding face to the latent space of the pretrained FFHQ model \cite{karras2020analyzing} and then iteratively optimize the face latent codes and body latent codes by the FFHQ \cite{karras2020analyzing} and our generator, respectively. Figure \ref{fig:insetgan} shows the results of combining geometry (Column \emph{Ours Body + Geometry Face}) or appearance (Column \emph{Ours Body + Appearance Face}) faces with bodies generated by our full-body generator. After optimizing both latent codes, we successfully obtain composition results that maintain coherence and preserve the identity.

\begin{figure}[h]
  \centering
  \includegraphics[width=0.9\linewidth]{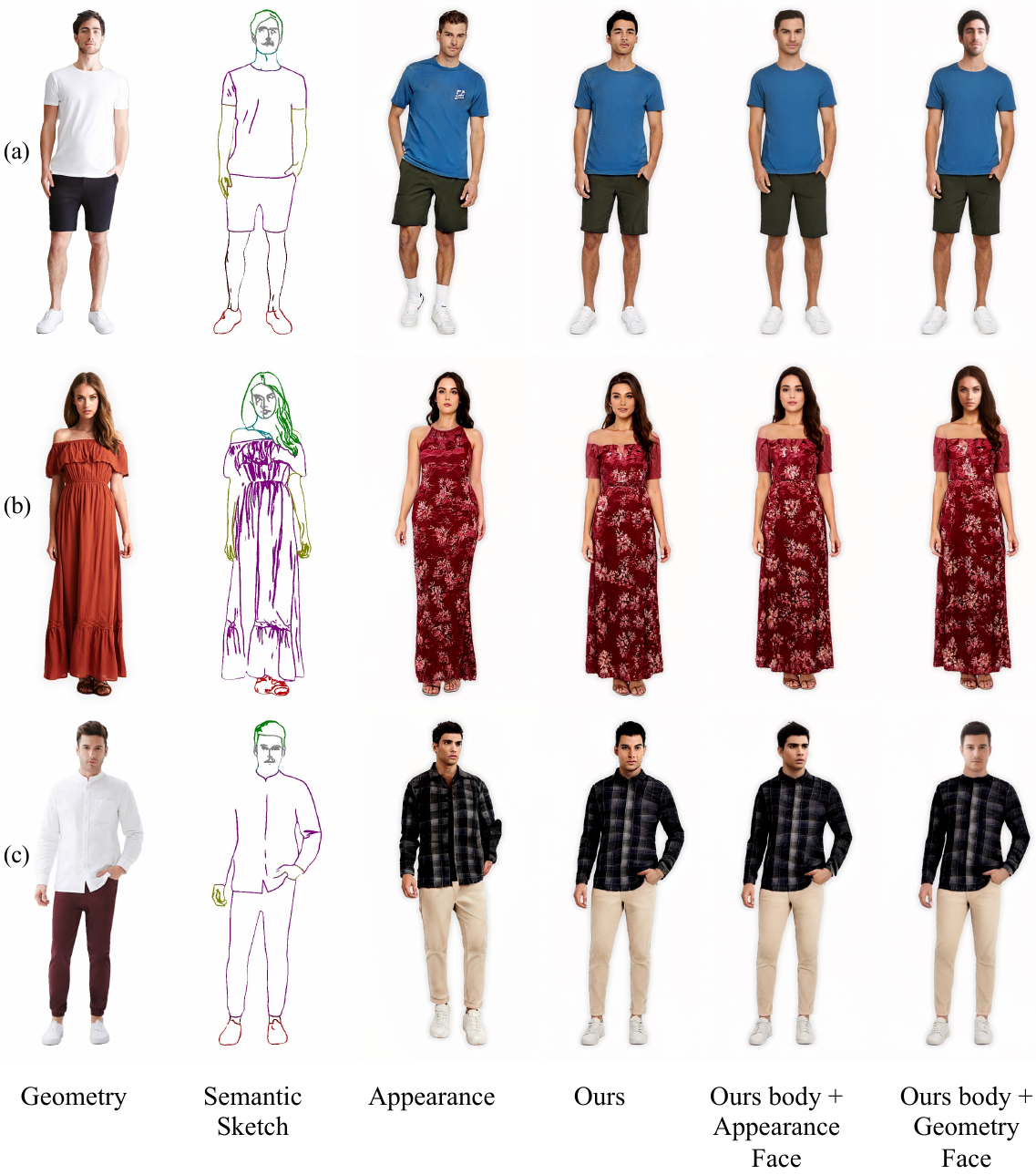}
  \caption{{Three face body montage results using InsetGAN. Given target geometry faces/appearance faces and bodies generated by our generator with the input of semantic sketches (extracted from geometry images) and appearance images, InsetGAN jointly optimizes the corresponding latent codes to achieve coherent results.}}
  \label{fig:insetgan}
\end{figure}

\subsection{Real Appearance Image}\label{sec:real}
{For a real appearance image, we first need to encode it into the latent space of StyleGAN-Human as $w_a$ using an off-the-shell encoder \cite{tov2021designing}. As Figure \ref{fig:real_img} shows, with the accurately inverted latent codes, our method successfully transfers the appearance of the real images. }

\begin{figure}[h]
  \includegraphics[width=\linewidth]{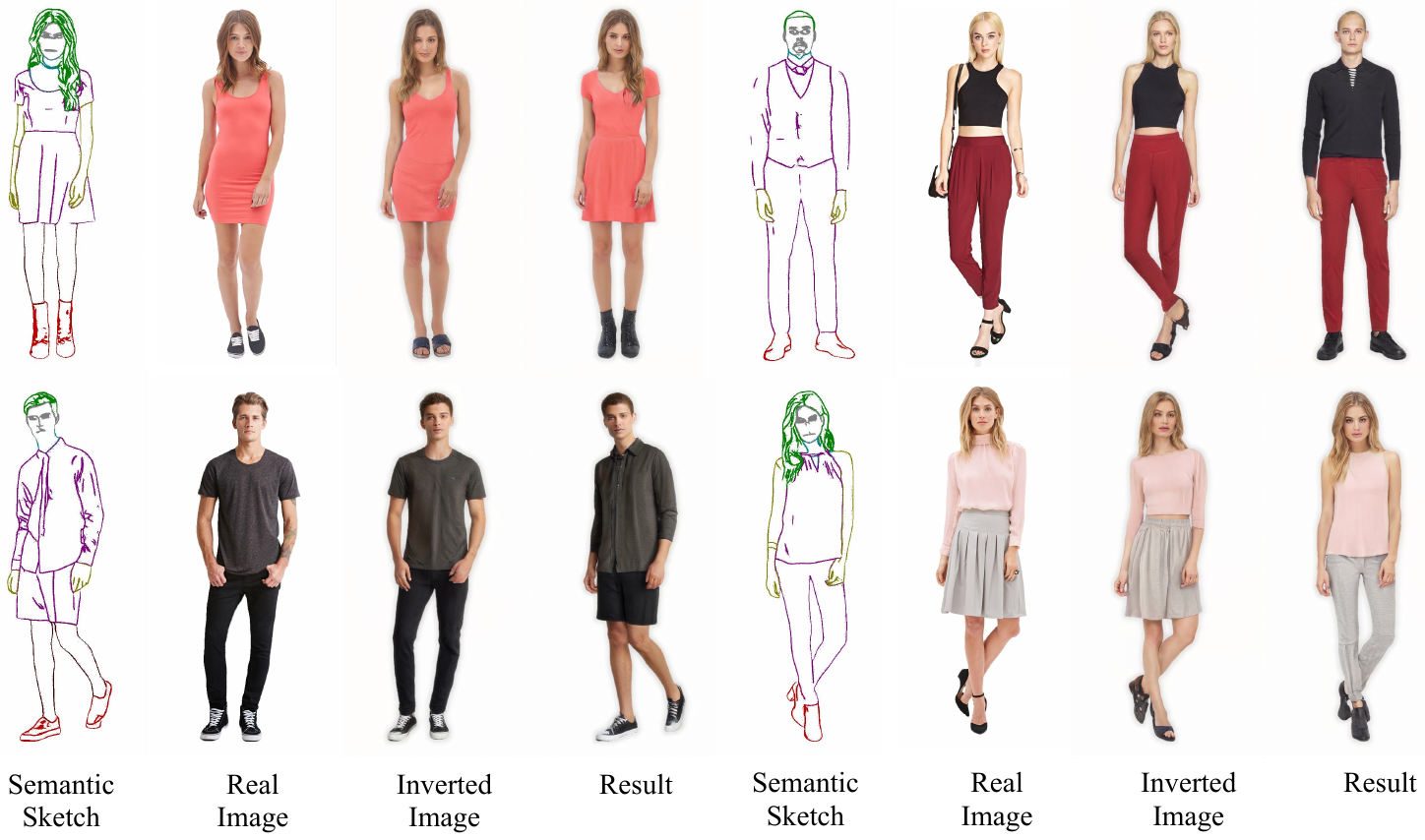}
  \caption{{The results with the inputs of a semantic sketch and a real appearance image. To achieve them, we first 
  invert the real appearance image to the latent space of StyleGAN-Human via an off-the-shell encoder \cite{tov2021designing}.}}
  \label{fig:real_img}
\end{figure}

\subsection{User Study}\label{sec:user}
We first prepared a set of input appearance images (50 in total) with diverse styles randomly picked from StyleGAN-Human, and then applied the four compared methods (Section \ref{sec:qual}) to each input appearance image with a semantic sketch image (25 randomly selected from our test set and 25 from freehand sketches (Supplementary Materials Section II)). The evaluation was done through an online questionnaire. There were, in total, 50 participants. We showed each participant all the compared results in random order with the sketch and appearance inputs, set by set. Noted that we exclude e4e since its input is an RGB image, but there is no such images for those user-drawn sketches. Tables \ref{tab:us_cm_1} and \ref{tab:us_cm_2} show the statistics of the voting results from the different style sketches. No matter what level of abstraction the input sketches were, our method was most chosen for every aspect. Therefore, human preference proves that with diverse inputs, our method maintains geometry and texture well and generates high-fidelity results.

\begin{table}[t]
\newcommand{\tabincell}[2]{\begin{tabular}{@{}#1@{}}#2\end{tabular}}
\begin{center}
 \begin{tabular}{ccccc}
 \hline
	\multicolumn{1}{c}{\textbf{Method}}
        & \multicolumn{1}{c}{\textbf{GP}$(\%)$} 
	& \multicolumn{1}{c}{\textbf{AT}$(\%)$} 
	& \multicolumn{1}{c}{\textbf{VQ}$(\%)$}
         & \multicolumn{1}{c}{\textbf{UP}$(\%)$}\\
	\hline
    PsP  & 5.2 & 7.3 & 6.6 & 6.3\\
    e4e  & 11.0 & 13.9 & 15.7 & 15.6\\
    CocosNetV2  & 31.6 & 5.9 & 6.9 & 7.6\\
    T2I-Adapter & 12.2 & 4.5 & 4.5 & 4.5\\
    \hline
    Ours  & \textbf{40.0} & \textbf{68.4} & \textbf{66.3} & \textbf{66.0}\\
    \hline
\end{tabular}
\end{center}
\caption{Summary of the voting results from the user study with the input sketches extracted from the DeepFashion dataset.}
\label{tab:us_cm_1}
\vspace{-2.5em}
\end{table}

\begin{table}[t]
\newcommand{\tabincell}[2]{\begin{tabular}{@{}#1@{}}#2\end{tabular}}
\begin{center}
 \begin{tabular}{ccccc}
 \hline
	\multicolumn{1}{c}{\textbf{Method}}
        & \multicolumn{1}{c}{\textbf{GP}$(\%)$} 
	& \multicolumn{1}{c}{\textbf{AT}$(\%)$} 
        & \multicolumn{1}{c}{\textbf{VQ}$(\%)$} 
	& \multicolumn{1}{c}{\textbf{UP}$(\%)$}\\
	\hline
    PsP  & 6.6 & 9.4 & 10.8 & 9.4 \\
    CocosNetV2  & 28.8 & 2.8 & 3.1 & 3.1\\
    T2I-Adapter  & 10.4 & 4.5 & 4.5 & 4.2\\
    \hline
    Ours  & \textbf{54.2} & \textbf{83.3} & \textbf{81.6} & \textbf{83.3}\\
    \hline
\end{tabular}
\end{center}
\caption{Summary of the voting results from the user study with the input sketches collected from users (Supplementary Materials Section II) via our interface.}
\label{tab:us_cm_2}
\vspace{-2.5em}
\end{table}

\subsection{Ablation Study}\label{sec:ab}
We report some results of the ablation study of our method in terms of the effectiveness of the Sketch Image Inversion module (Section \ref{sec:sketch}) and the Body Generator Tuning module (Section \ref{sec:body}). We use semantic sketches as input for the ablation study. 

\subsubsection{Effectiveness of Sketch Image Inversion}\label{sec:ab1}
Since the StyleGAN-Human cannot fully capture the distribution of complex real full-body images, the encoder supervised with those real images results in inaccurate geometry and artifacts (Figures \ref{fig:cm_sketch} and \ref{fig:cm_user}). We adopt the sampled data to train our Sketch Image Inversion module with a series of loss functions. We evaluate the effectiveness of each loss by testing the performance of our model trained in various ways, including trained only with the full supervised loss ($w/ L_{l_2}$), trained with the self-supervised loss ($w/ L_{lpips}$), trained with both of them ($w/ L_{l_2} + L_{lpips}$). All the experiments are implemented with $L_{adv_w}$.

As shown in Figure \ref{fig:ab_encoder}, only the results supervised with all the proposed losses (\emph{ours}) achieve consistent geometry (e.g., poses, clothing types) with the sketch inputs. While the model trained with $L_{l_2}$ can synthesize high-quality results (Column $w/ L_{l_2}$), these results do not faithfully respect the input sketches. For example, cases ((a) and (e)) change the head pose, and cases ((b) and (c)) produce undesirable clothing contours. It is mainly because full supervision can stably predict the geometric code conforming to the distribution, thus ensuring the generation quality, but small deviations in the latent space would be amplified and reflected in the geometric shape. The encoder supervised with $L_{lpips}$ provides more accurate poses and shapes (Column $w/ L_{lpips}$) but introduces artifacts on the clothing ((a) and (b)) and influences the color representative ((c)), possibly due to the strong geometry constraint to ignore the original distribution. Hence, applying both of them can solve the above issues to a certain extent (Column $w/ L_{l_2} + L_{lpips}$). However, due to the sparsity of sketches, some line segments can easily be ignored at the training stage. That leads to the unstable prediction of semantic elements ((a)), especially for those (e.g., glasses, hats) with a low probability of occurrence 
in the training data and inaccurate clothing boundaries, especially when clothing and body contour tend to be consistent ((c)). To solve this problem, we explicitly add the semantic supervision $L_{ce}$ and $L_{dice}$ so that the encoder can focus on the accuracy of semantic objects separately.
\begin{figure*}[t]
  \includegraphics[width=0.9\textwidth]{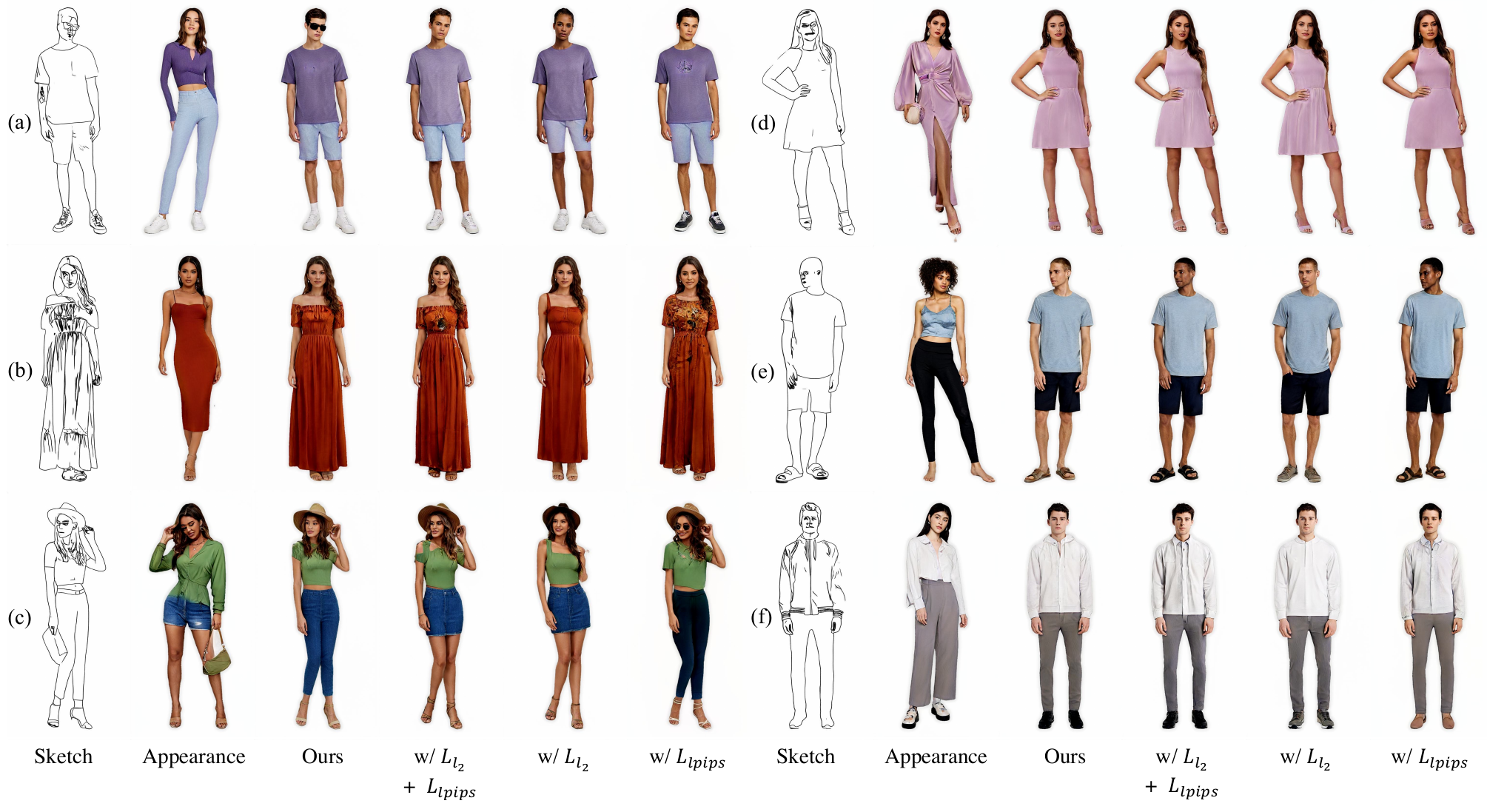}
  \caption{The ablation study for the Sketch Image Inversion module.}
  \label{fig:ab_encoder}
\end{figure*}

According to the quantitative metrics provided in Section \ref{sec:qua_cm}, we show the quantitative results of the ablation study for the Sketch Image Inversion module in Table \ref{tab:qua_encoder}.
The baseline model with $L_{lpips}$ gets the highest CD. This is because the artifacts inside the generated images can heavily influence the CD metric. Our full model achieves the best performance on the MIoU metric. Meanwhile, removing any components degrades the model's performance integrally.

\begin{table}[t]
\begin{center}
 \begin{tabular}{ccc}
 \hline
	\multicolumn{1}{c}{\textbf{Method}}
	& \multicolumn{1}{c}{\textbf{CD $\downarrow$}} 
        & \multicolumn{1}{c}{\textbf{MIoU $\uparrow$}}\\ 
	\hline
    $w/ L_{l_2}$  & 7.06 & 0.68 \\
    $w/ L_{lpips}$ & \textbf{3.95} & 0.75 \\
    $w/ L_{l_2} + L_{lpips}$ & 4.98 & 0.77 \\
    \hline
    Ours  & 4.95 & \textbf{0.80} \\
    \hline
\end{tabular}
\end{center}
\caption{Quantitative results of the ablation study for the Sketch Image Inversion module on the DeepFashion dataset.}
\label{tab:qua_encoder}
\end{table}

\subsubsection{Effectiveness of Body Generator Tuning}\label{sec:ab2}
Since the high-resolution layers of the existing StyleHuman-GAN fail to control the specific textures, we propose the Body Generator Tuning module to alter the weights. To validate the efficiency of this module, we compare our method with its variants from the perspective of loss terms, including supervised with a single appearance input ($w/ I_a$), supervised with multiple appearance images after mixing at layer 6 ($w/ I_{mix6}$), and supervised with multiple appearance images after mixing at layer 6 and the semantic loss  ($w/ I_{mix6} + L_{dice}$). We only fine-tune the high-resolution layers for the above experiments. We also implement the full-weight updates (Full-layer).

As shown in Figure \ref{fig:ab_decoder}, it is obvious that supervised with a 
single appearance input, the results exhibit mean textures but lack the global and local structure distribution (Column $w/ I_a$). Those textures do not vary with 
the poses of the input sketches ((c) and (e)). Additionally, for cases with multiple textures of different clothing, the results tend to mix textures ((b) and (e)). The possible reason is that the optimization with a single appearance image easily falls into a local optimum.
Therefore, we propose to fine-tune the generator with the style-mixing results $I_{mix6}$, thus improving the quality of textures (Column {$w/ I_{mix6}$}). However, the body and clothing shapes change a lot ((b) and (f)) since $I_{mix6}$ is not completely aligned with the geometric requirements. By comparing Columns $w/ I_a$ and $w/ I_{mix6} + L_{dice}$, we can see that this issue is greatly solved after incorporating semantic supervision. 
However, such semantic supervision leads to the emergence of non-photorealistic faces (see (b) and (f) in Column {$w/ I_{mix6} + L_{dice}$}). In other words, this supervision affects the fidelity of the generator. As shown in Column \emph{Ours}, by adding the content loss with $I_{mix10}$, the model avoids such artifacts. Although there is a slight difference in the outer contour between full-layer updates and high-resolution-layer updates (see (a) and (e) in Column \emph{Full-layer}), only fine-tuning the high-resolution layers can decrease the number of training parameters, thus slightly reducing the time required for fine-tuning.

\begin{figure*}
  \includegraphics[width=\textwidth]{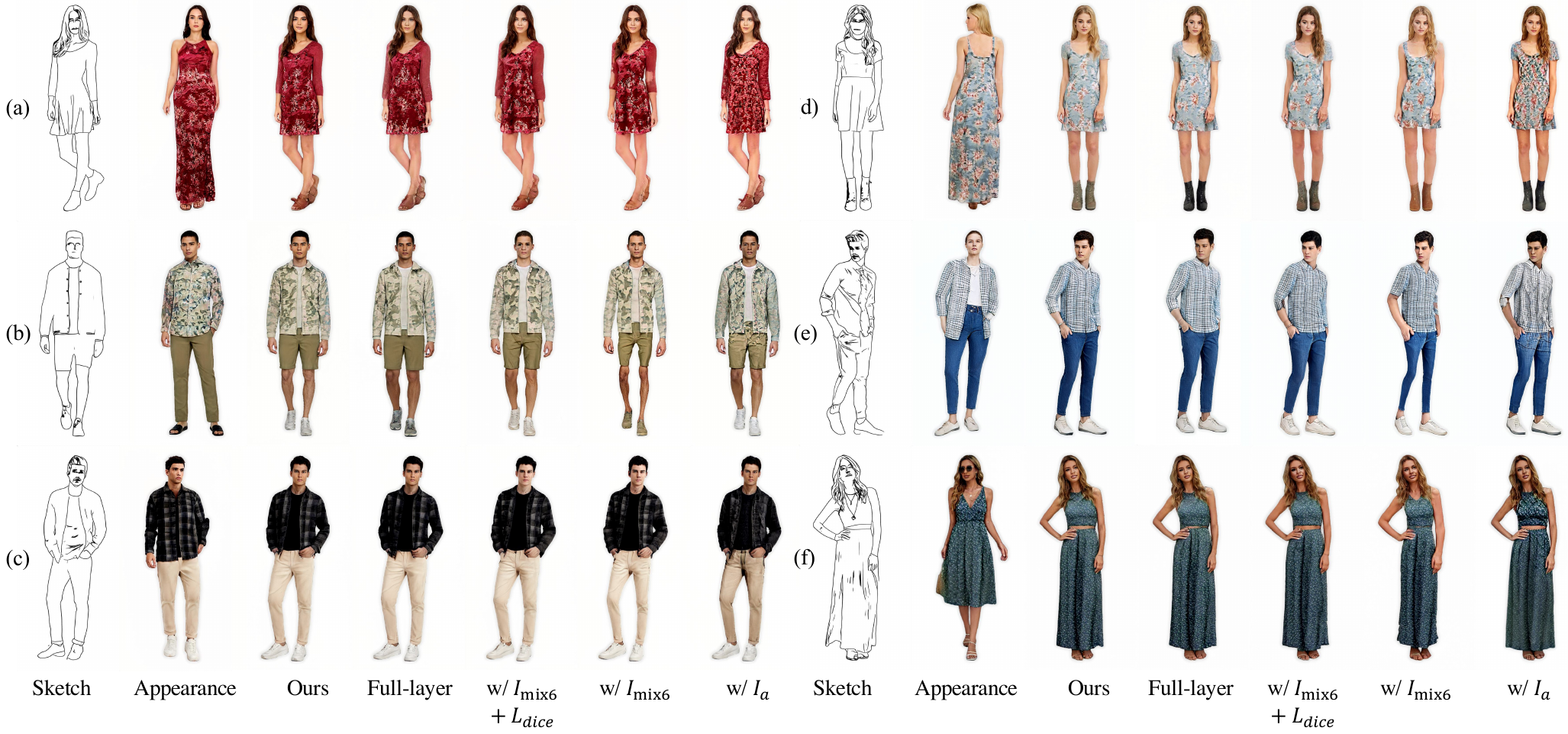}
  \caption{The ablation study for the Body Generator Tuning module.}
  \label{fig:ab_decoder}
\end{figure*}


\section{Conclusion and Limitations}
We presented~\sysName, the first system for controllable human full-body image generation given sketch and appearance constraints. The first stage of our system inverts the sketch input to the latent space. To achieve high-precision geometric encoding, we proposed to train our sketch encoder with infinitely sampled images and directly calculate the loss on the sketch level. Its second stage aims to enhance the texture expressiveness for the existing StyleGAN-Human under different postures and shapes. Due to the lack of a proper dataset, we proposed two style-mixing strategies to synthesize the data that well preserve the texture and geometric information, respectively. Although they are changed on the other side, they can be used as good guidance together. Hence, we utilized these results to fine-tune the generator. Extensive experiments have demonstrated that our \sysName~ generates results preserving both geometry and texture/color consistency with the two inputs and outperforming the four compared methods. 
The flexible control of inputs and attractive results from the applications (Supplementary Materials Section IV) show the practicality and significance of our method.

Our method can still be improved in various ways. First, since we choose to embed the sketch input into the latent space, our method prefers to produce reasonable results but sometimes ignores the user's intent (Figure \ref{fig:failure} (b)). This problem could be solved by embedding the spatial guidance for the generator. Second, for real appearance images, our method is influenced by the appearance codes from the image inversion method \cite{tov2021designing}. For those with complex textures, the corresponding textures cannot be transferred due to the inaccurate appearance codes and the limited generative power of StyleGAN-Human (Figure \ref{fig:failure} (a)). Additionally, our method relies on the style mixing results of StyleGAN-Human. Such results fail to keep the hairstyles stable and miss the appearance of the shoes and glasses. These issues might be alleviated or addressed by more powerful full-body StyleGAN. 

\begin{figure}[h]
  \centering
  \includegraphics[width=\linewidth]{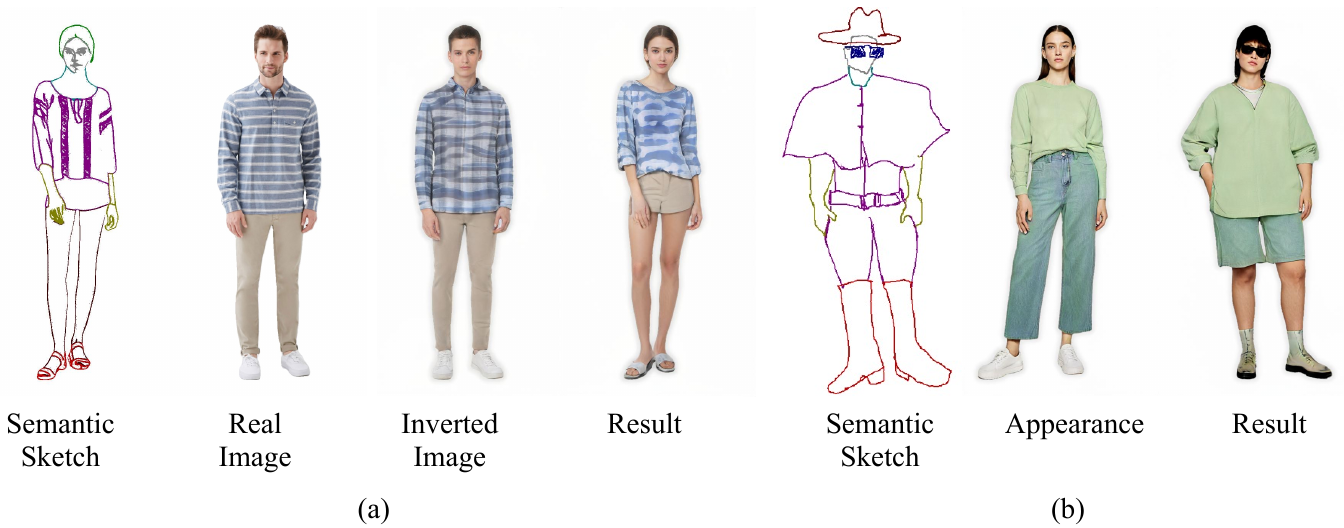}
  \caption{Two less successful cases. (a) covers an example for transferring the appearance of a real image with e4e inversion. (b) shows an unsatisfactory result, which respects the training image distribution but not the creative design.}
  \label{fig:failure}
\end{figure}

\bibliographystyle{IEEEtran}
\bibliography{BIB_all}

\end{document}